\documentclass[11pt,a4paper]{article}
\usepackage[hyperref]{acl2021}
\usepackage{times}
\usepackage{latexsym}
\usepackage{amsmath}

\usepackage{linguex}

\usepackage{setspace}
\usepackage{graphicx} 

\usepackage{etoolbox}
\makeatletter
\pretocmd{\@ex@word}{\normalfont}{}{}
\pretocmd{\@gl@word}{\normalfont}{}{}
\makeatother

\usepackage{microtype}

\title{Interpretable dimensions support an effect of agentivity and telicity \\on split intransitivity}

\author{Eva Neu \and Brian Dillon \and Katrin Erk \\
  University of Massachusetts, Amherst \\
  \texttt{eneu,bwdillon,kerk@umass.edu}} 
 
\date{}

\begin{document}

\maketitle
\begin{abstract}

Intransitive verbs fall into two different syntactic classes, unergatives and unaccusatives. It has long been argued that verbs describing an agentive action are more likely to appear in an unergative syntax, and those describing a telic event to appear in an unaccusative syntax. However, recent work by \citet{KimDecomposing} found that human ratings for agentivity and telicity were a poor predictor of the syntactic behavior of intransitives. Here we revisit this question using interpretable dimensions, computed from seed words on opposite poles of the agentive and telic scales. Our findings support the link between unergativity/unaccusativity and agentivity/telicity, and demonstrate that using interpretable dimensions in conjunction with human judgments can offer valuable evidence for semantic properties that are not easily evaluated in rating tasks.

\end{abstract}

\section{Introduction}

It has long been known that the syntactic structures in which a verb can appear depend on its lexical semantics \citep{Dowty, Jackendoff83, LevinEnglish, LRH, VanValin}. For instance, verbs denoting a transfer of possession such as \textit{give} typically appear in a ditransitive syntax, and verbs denoting a causal relation such as \textit{kill} in a transitive syntax. 

One area in which the effect of lexical semantics on syntax has been discussed most extensively is split intransitivity. A wide variety of syntactic constructions distinguish between two kinds of intransitives. For instance, \textit{freeze} but not \textit{play} can surface in prenominal participle constructions \ref{diagnostic}:

\ex. \label{diagnostic} \a. \label{opened}the frozen lake
\b. \label{laughed}*the played child

\noindent Moreover, \textit{freeze} but not \textit{play} can take a secondary predicate describing the result of the event \ref{resultative}:

\ex. \label{resultative}\a. The lake froze solid. 
\b. *The child played tired. (i.e., became tired as the result of playing)

\noindent The fact that these two -- and other -- syntactic diagnostics converge has led researchers to posit two different syntactic structures for intransitives, unergatives and unaccusatives \cite{BurzioDiss, BurzioBook, Perlmutter}. Unaccusative structures allow for prenominal participle constructions and resultative predicates; unergative structures do not. In \ref{diagnostic} and \ref{resultative}, \textit{freeze} behaves as an unaccusative and \textit{play} as an unergative. 

How likely verbs are to show unergative or unaccusative behavior has been linked to their lexical semantics. Verbs denoting a strongly agentive activity have been argued to be unergative-leaning, and verbs denoting a telic event to be unaccusative-leaning \citep{SoraceLang, SoraceArchivio, SoracePuzzle}. Agentivity is an umbrella term for properties such as sentience, volition, purpose and causal power \citep{Dowty}. Telicity describes events that have a natural endpoint and result in a particular state \citep{RothsteinEvents, TennyDiss, TennyBook, Vendler}. For instance, the verb \textit{play} denotes a more agentive activity than (intransitive) \textit{freeze}, and \textit{freeze} but not \textit{play} describes a telic event in which the argument undergoes a change of state.

While this correlation between agentivity and unergativity, and between telicity and unaccusativity is a staple of the literature, it also remains controversial and poorly understood. Some have denied it altogether \citep{Graf}. Others have argued that its validity is limited and that other semantic factors play a larger role \citep{KimDecomposing}. A few computational and experimental studies \citep{AcarturkZeyrep, AllmannUnacc, BakerSplit, HuangUnaccusativity} have found some support for the correlation, but are subject to various limitations. Moreover, there is no agreement on the precise definition of agentivity and telicity, especially whether they should be understood as categorical or gradient properties \citep{KimDecomposing}. Most importantly, from a methodological angle, it is not at all clear how to evaluate whether or to what extent a verb qualifies as agentive or telic. 

In the present paper, we investigate the semantic correlates of the unergative/unaccusative distinction using word type embeddings.  We set up interpretable dimensions for agentivity and telicity in vector space by specifying for each of them a set of positive and a set of negative seed words and averaging over the difference vectors. The projection of a word embedding on this axis encodes to what extent the word is associated with the positive or the negative end of this semantic dimension. Seed-based dimensions have in the past been used for semantic properties such as danger, size \citep{GrandDim} and affluence \citep{KozlowskiDim}. The similarities derived from these dimensions have been shown to match human judgments to a considerable extent, indicating that they  encode rich conceptual knowledge. 

We evaluate our interpretable dimensions for agentivity and telicity by comparing them to a syntactic measure of unergativity/unaccusativity, taken from \citet{KimDecomposing}, as well as to human ratings for agentivity, telicity and animacy. In previous work on interpretable dimensions, human judgments have been used as the gold standard against which the performance of interpretable dimensions is measured. Here we choose a different strategy. Agentivity and telicity are complex and abstract notions that are sensitive to context. There is no consensus on how to collect ratings for them. Instead, we set up property axes so as to maximize the fit to the syntactic measure of unergativity/unaccusativity, and then use human ratings to better understand what these axes correspond to. In this context, we also develop a new method for fitting dimensions directly to human ratings that does not rely on seed words.

On a theoretical level, we find support for a correlation between agentivity/telicity and the unergative/unaccusative distinction. On a methodological level, we explore new ways of using human judgments and word embeddings in conjunction to better understand semantic properties that are not easily amenable to a regular rating task.

\section{Previous work on the semantic correlates of split intransitivity}

The Unaccusativity Hypothesis developed in the generative tradition holds that the unergative/unaccusative distinction corresponds to two different underlying syntactic structures (\citealt{BurzioDiss, BurzioBook, Perlmutter}, but see \citealt{VanValin}). In a nutshell, in unergatives, the sole argument has the same syntactic status as the subject of transitives; in unaccusatives, it has the same status as the object of transitives. 

Under this analysis, the semantic correlates of the unergative/unaccusative split have their roots in $\theta$-theory. Researchers such as \citet{Dowty} have argued that the subject argument of transitives receives an agent $\theta$-role, associated with properties such as sentience, volition, purpose and causal power, and that the object argument receives a patient $\theta$-role associated with properties such as undergoing a change of state.  If the arguments of unergatives and unaccusatives have the same syntactic status as the subject and object arguments of transitives, this can be taken to predict that they are equally associated with agentivity and with undergoing a change of state, respectively.


This correlation between unergativity and agentivity, and between unaccusativity and telicity has been defended most strongly by  \citet{SoraceLang, SoraceArchivio, SoracePuzzle}.
She proposed that intransitive verbs can be categorized into seven different classes, such as `change of state' or `uncontrolled process,' which sit along a spectrum ranging from atelic and strongly agentive to telic and weakly agentive. Sorace argued that verbs at the atelic/agentive end of the spectrum behave as unergatives and those at the telic/non-agentive end as unaccusatives, with variable behavior verbs in between. In addition, agentivity and telicity also predict different syntactic behavior for different tokens of the same verb. In Dutch, unergatives take the auxiliary \textit{have}, and unaccusatives \textit{be}. While \textit{roll} in isolation describes an atelic event realized with an unergative syntax, \textit{rolling downstairs} describes a telic event requiring an unaccusative syntax  \ref{aux}.

\ex. \label{aux}\ag. \normalsize{De} \normalsize{bal} \normalsize{\textit{heeft}} \normalsize{gerold}.\\
\normalsize{the} \normalsize{ball} \normalsize{\textit{has}} \normalsize{rolled}\\
`The ball rolled.' \label{aux1}
\bg. \normalsize{De} \normalsize{bal} \normalsize{\textit{is}} \normalsize{naar} \normalsize{beneden} \normalsize{gerold}.\\
\normalsize{the} \normalsize{ball} \normalsize{\textit{is}} \normalsize{to} \normalsize{down} \normalsize{rolled}\\
`The ball rolled downstairs.'\label{aux2}

\vspace{-5mm}

\begin{flushright}
    (\citealt{SoraceLang}:876)
\end{flushright}

More recent work has attempted to find large-scale empirical support for Sorace's claims. \citet{AcarturkZeyrep} found that feedforward neural networks trained either on semantic features such as telicity and volitionality or on syntactic acceptability judgments for unaccusativity diagnostics largely succeed in dividing verbs into unergatives and unaccusatives, and also varied in their predictions for Sorace's variable behavior verbs. Experimental studies on unaccusativity diagnostics have consistently found that verbs cannot be neatly divided into unergatives and unaccusatives, but only some found support for Sorace's verb classes \citep{AllmannUnacc, BakerSplit, HuangUnaccusativity}. However, \citet{KimDecomposing} have pointed out that these studies suffer from various limitations, such as including only a limited number of verbs and presupposing that syntactic judgments and semantic features are binary.

Therefore, \citet{KimDecomposing} conducted a systematic comparison of various semantic measures to determine their predictive value for the syntactic behavior of English intransitives. These measures include Sorace's seven verb classes as well as a similar verb classification developed by \citet{LevinEnglish}. Another set of predictors was derived from the GloVe embeddings of the verbs by means of a Principal Component Analysis. In addition, Kim et al. collected ratings for each verb  on a scale from 0 to 6 for two sets of features. One consisted of 6 event-related features traditionally considered relevant for the unergative/unaccusative distinction (Agentivity, Telicity, Caused, Transitivity, Dynamicity, Requires energy input). The other contained 66 properties developed by \citet{BinderFeatures} to capture how concepts are represented through high-level brain-based features, such as Color, Pain, Duration and Angry.

To evaluate the performance of these predictors, Kim et al. collected ratings for 138 verbs on how acceptable they are in prenominal participle constructions, a widely used unaccusativity diagnostic. Ratings are expected to be higher for more strongly unaccusative verbs. Each verb was presented in the context of three different phrases and rated on a 1--5 Likert scale. The event-related features predicted the syntactic data better than Levin's and Sorace's categorizations, but only Caused emerged as a significant predictor after correcting for multiple comparisons. The model combining experiential and event-related features predicted the syntactic ratings  best; Caused and Agentivity were significant predictors among several others. The GloVe-based model came in second. Kim et al. conclude that the unergative/unaccusative distinction is rooted in graded, embodied features of sensory experience.

Kim et al.'s findings suggest that the effect of agentivity and telicity on split intransitivity might have been overestimated. Telicity in particular was surprisingly not a significant predictor in any of their models. However, Kim et al. operationalized agentivity and telicity using human judgments. This introduces two potential confounds. First, the questions subjects were asked might not have targeted exactly the right concepts. Second, agentivity and telicity might be difficult to evaluate by subjects in a rating task. In particular, Kim et al. collected ratings for verbs in isolation, which could distort intuitions. Hence, we investigate agentivity and telicity using interpretable dimensions, which are less susceptible to these confounds.

\section{Interpretable axes in embedding space}

Many gradable properties seem to be linearly encoded in embedding space, a fact that has been used for analyses in linguistics, cognition, and social sciences~\citep{BolukbasiDim, KozlowskiDim,gari-soler-apidianaki-2020-bert,GrandDim}. It has even been proposed that many high-level concepts are encoded linearly~\citep{pmlr-v235-park24c}, although this does not hold for all concepts~\citep{Engels-nonlinear}.

\paragraph{Seed-based axes.} A simple and widely used method for obtaining encoding axes for gradable properties is through manually defined seed words. For example, for \textit{danger}, the two poles of the property could be described with 
$\{$\textit{dangerous}, \textit{unsafe}$\}$ and $\{$\textit{safe}, \textit{harmless}$\}$, respectively. An axis for \textit{danger} can then be obtained as the mean over all difference vectors 
$\vec{\text{\textit{safe}}} - \vec{\text{\textit{dangerous}}}$, 
$\vec{\text{\textit{safe}}} - \vec{\text{\textit{unsafe}}}$, 
$\vec{\text{\textit{harmless}}} - \vec{\text{\textit{dangerous}}}$, 
$\vec{\text{\textit{harmless}}} - \vec{\text{\textit{unsafe}}}$. 

Say $\vec f$ is an axis for property $f$, then the degree or rating of a word $w$ on $f$ is predicted to be the scalar projection onto the axis: 
\[ ||\text{proj}_{\vec w}(\vec f)|| = \frac{\vec w \cdot \vec f}{||\vec f||}\]

\paragraph{Fitted axes.}
\citet{ErkApidianikiAdjust} (below: EA) introduce a method for computing property axes that interpolates seed words with training data in the form of human ratings. To compute an axis $\vec f$ for a property $f$ of concepts $w$, they use a loss function that penalizes the predicted scalar projection of $\vec w$ onto $\vec f$ for deviation from the gold rating $y_w$ for $w$. This is combined with a second loss that enforces closeness to a seed-based axis. 

The EA model optimizes pointwise fit to human ratings. But the typical use of property axes is to predict \emph{rankings} of concepts along the property. We introduce a new model that, like the EA model, fits a property axis to human ratings, but uses a ranking loss to more directly approximate the characteristic of interest. We use a margin ranking loss~\citep{nayyeri2019adaptivemarginrankingloss}: for any pair $(a, b)$ of concepts where the gold rating of $b$ is higher than that of $a$, it encourages the predicted value for $b$ to be higher than $a$'s by at least a margin of $d$:
$$J_r = \sum_{(a, b) \in P, \hat y_b > \hat y_a|} max(0, d - y_b + y_a)$$
where $P$ is a set of training item pairs, an $N$-size set sampled from all training item pairs with at least a difference of $d$ in their gold ratings; for a concept pair (a, b), $\hat y_a, \hat y_b$ are gold ratings, and $y_a, y_b$ are predictions. $N$ and $d$ are the parameters of the model. 

We evaluate the ranking model on the large collection of concepts and human property ratings introduced by \citet{GrandDim} and compare to the original seed based axes of \citet{GrandDim} and the EA model. \citet{GrandDim} measure performance through correlation as well as \emph{pairwise order consistency (poc)}, the percentage of test pairs ordered correctly by the model. However, we cannot use correlation because the data sets become too small once part of the data is used for training. We measure poc as well as \emph{extended poc (xpoc)}, the percentage of test pairs and train/test pairs ordered correctly by the model~\citep{ErkApidianikiAdjust} -- this checks whether test items are ranked  correctly with respect to training or test items. Results on 5-fold crossvalidation on the Grand et al.\ data are shown in Table 1. \textit{Ranking} is our new model. In the evaluation, as throughout in this paper, we use GLoVE embeddings with 300 dimensions pre-trained on Wikipedia and Gigaword. 
\footnote{Hyperparameters of the EA model are as given in that paper; hyperparameters for our ranking model were optimized on the development portion of the Grand
et al. (2022) data defined by Erk and Apidianaki (2024), $d = 0.2 \cdot $ sdev, $N = 300$.}

\begin{table}
    \centering
    \begin{tabular}{lccc}
        model & poc & xpoc\\\hline
     \rule{0pt}{12pt}Seed    & .629 & .631 \\
     Pointwise    &  .701&  .779\\
     Ranking    & .703 & .797 \\
    \end{tabular}
    \label{tab:grandranking}
    \caption{Comparing our ranking-based fitted axes to EA (Pointwise) and seed-based axes on the data of \citet{GrandDim}. Pointwise uses interpolated losses based on human ratings and seed axes.}
\end{table}

We see that the ranking loss model, like the EA model, clearly improves fit over the seed-based axes, and achieves even better performance than the EA model. The EA model interpolates human ratings with seeds, and in fact flounders when no seed axis is given. The ranking model, in contrast, manages to fit the data very well even without the help of a seed axis.

Fitted dimensions thus allow us to use existing data sets of human ratings to extrapolate simulated ratings for unseen items. In this paper, we later employ fitted dimensions to simulate animacy ratings for intransitive verbs.

\section{Modeling}

We have argued that human ratings for agentivity and telicity are subject to potential pitfalls, which might obscure their correlation with syntactic unergativity/unaccusativity. Instead, we set up interpretable dimensions for these properties that optimize the fit to the syntactic data and then compare them to human judgments. Our strategy is to find the dimensions in space that syntax appears to be sensitive to and then analyze which semantic properties precisely they encode.

We set up seed-based interpretable dimensions and then compute for Kim et al.'s intransitives where they fall on these axes. To determine the best set of seeds, we tested different words for whether they improve the fit of a regression model predicting Kim et al.'s syntactic data, i.e., acceptability ratings of the different verbs in prenominal participle constructions. Specifically, we ran simple linear regression models with only the seed-based predictors and observed whether adding or removing a seed word improved the model's R\textsuperscript{2} score.

The best seeds we found for agentivity were  \{\textit{think}, \textit{you}, \textit{he}, \textit{she}, \textit{causally}\} (positive) and \{\textit{affected}\} (negative). For telicity, they were  \{\textit{result}, \textit{effect}, \textit{completely}, \textit{fully}, \textit{eventually}\} (positive) and \{\textit{still}, \textit{ongoing}, \textit{being}, \textit{acting}\} (negative). Several words associated with agentivity and telicity, respectively, did not improve the model's fit and were not included in the final set of seeds. These included, for agentivity, \textit{agent}, \textit{deliberate}, \textit{intentional}, \textit{purpose}, \textit{volition} and \textit{active} (positive) and \textit{patient}, \textit{stationary} and \textit{it} (negative). For telicity, words like \textit{outcome}, \textit{complete} and \textit{entirely} (positive) and \textit{continuous}, \textit{open-ended} and \textit{unfolding} (negative) did not make the cut. We discuss later why certain words but not others turned out to improve the model.

\section{Comparing dimensions and ratings}

To verify that the seed based agentivity and telicity dimensions predict the syntactic data well, we fitted mixed-effects Bayesian ordinal regression models with default priors, using the \texttt{brms} library in R. For comparison, we also fitted the same models using Kim et al.'s agentivity and telicity ratings as predictors. Both models were computed with a cumulative probit link function and fitted with 2000 iterations (1000 warm-up, 1000 samples taken). R-hat was 1.00 throughout; no divergences were observed during sampling. All gradable features were z-scored. Unlike \citet{KimDecomposing}, we did not average the syntactic ratings over phrases and subjects, which would not allow us to account for inter-subject variation and would generally simplify the data, potentially obscuring important effects. Instead, all our models included by-subject intercepts.

Table \ref{reg} summarizes the regression coefficients for the seeds and the ratings model. As expected, in both models, agentivity decreases and telicity increases the acceptability of prenominal participle constructions, an unaccusativity diagnostic.

\begin{table}[h!]
    \centering
    \begin{tabular}{l c c c c }
    & Est. & Est.error & l-95\% & u-95\%\\
    \hline 
    \rule{0pt}{12pt}seeds ag. & -.24 & .01 & .27 & -.22  \\
    seeds tel. & .30 & .01 & .28 & .33\\
    \hline 
   \rule{0pt}{12pt}ratings ag. & -.21 & .01 & -.23 & -.19\\
    ratings tel. & .15 & .01& .13 & .17\\
    \end{tabular}
    \caption{Estimates of regression coefficients for seeds and ratings models, with estimated standard error and left and right boundary of the 95\% confidence interval.}
    \label{reg}
\end{table}

Figures 1 and 2 show the ratings predicted by each model plotted against the observed ratings. The seeds model clearly achieves a superior fit. 

\vspace{2mm}

\begin{minipage}{.45\columnwidth}

\includegraphics[scale=.52]{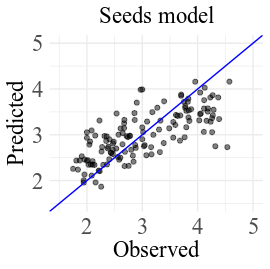}

\end{minipage}
\begin{minipage}{.45\columnwidth}

\includegraphics[scale=.52]{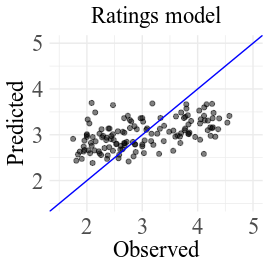}
    
\end{minipage}

\vspace{3mm}

\begingroup
\fontsize{10pt}{12pt}\selectfont
\noindent Figures 1 and 2: Observed and predicted mean ratings for each verb type. Predictions are derived from the posterior distributions of the seeds and ratings models.
\par
\endgroup

\vspace{4mm}

We further compared the goodness of fit of the seeds model and the ratings model with a leave-one-out (LOO) analysis using the \texttt{loo} library in R. LOO provides a measure of predictive accuracy by training a model on all data points except one and then testing how well the model predicts the held-out data point. Both the seeds model and the ratings model are compared against a null model containing only by-subject intercepts and a model that contains both seeds and ratings.\footnote{The formulas of the four models are: 

\noindent \texttt{answer $\sim$ (1|subj)}

\noindent \texttt{answer $\sim$ seeds\_ag+seeds\_tel+(1|subj)}

\noindent \texttt{answer $\sim$ rate\_ag+rate\_tel+(1|subj)}

\noindent \texttt{answer $\sim$ seeds\_ag+seeds\_tel+rate\_ag +} 

\texttt{rate\_tel+(1|subj)}} This allows us to evaluate how well the two semantic measures predict the syntactic data both in isolation and in conjunction with each other. 

Table \ref{loo} summarizes the result of the LOO analysis in terms of expected log predictive density (ELPD). ELPD is a measure of the log probability that the model attributes to all the held-out data points; a higher value (closer to zero) corresponds to better predictive performance. The seed-based predictor improved the ELPD considerably more than the rating-based predictor (755.4 vs. 232.2). Moreover, adding seeds to the ratings model vastly improves the fit (525.9), while there is no evidence that ratings improve on the seeds model (3.8). 

\begin{table}[h!]
    \centering
    \begin{tabular}{l c c}
         models &   ELPD diff &  SD \\
         \hline 
       \rule{0pt}{12pt}seeds vs. null model & 755.4 & 36.3\\
       ratings vs. null model & 233.2 & 21.8\\
       seeds+ratings vs. seeds & 3.8 & 3.6 \\
       seeds+ratings vs. ratings & 525.9 & 31.4\\
    \end{tabular}
    \caption{LOO analysis, showing by how much the first model improves the ELPD compared to the second, and  the standard deviation of this difference.}
    \label{loo}
\end{table}

\noindent In sum, seeds and Ratings, while both designed to capture agentivity and telicity, differ substantially in their predictive value for the syntactic ratings. The effect of the seeds subsumes the effect of the ratings, but also to encompasses something else. We now look at agentivity and telicity separately.


\paragraph{Telicity}

We performed another LOO comparison for models that only included telicity predictors \ref{loo_tel}. Again, the seeds model performs better than the ratings model, and adding ratings to the seeds model resulted in only small improvements.

\begin{table}[h!]
    \centering
    \begin{tabular}{l c c}
        models &  ELPD diff & SD \\
         \hline 
       \rule{0pt}{12pt}seeds vs. null model & 575.2 & 31.5\\
       ratings vs. null model & 69.2 & 11.8\\
       seeds+ratings vs. seeds & 1.2 & .6 \\
       seeds+ratings vs. ratings & 504.9 & 3.2\\
    \end{tabular}
    \caption{LOO analysis with only telicity predictors.}
    \label{loo_tel}
\end{table}

\noindent \citet{KimDecomposing} report high inter-speaker variation for the telicity ratings (the second highest among all features, SD = 1.8). We also assessed the test-retest reliability of the two seed-based and two rating-based predictors using the model-based analysis from \citet{StaubTestRetest}. For each set of predictors (seeds and ratings), we split up the data into even and odd tokens and fitted an ordinal Bayesian mixed-effects model to each of them. We then estimated the correlation between the subject-wise intercepts in each of the models to test the consistency of the fixed effects for each subject. We found that the telicity ratings have potentially low test-retest reliability (correlation .506, 95\% CI -.288--.956), considerably lower than agentivity ratings (.696, 95\% CI .146--.977), agentivity seeds (.955, 95\% CI .856--.997) and telicity seeds (.892, 95\% CI .750--.982). These findings signal the presence of measurement errors. 

Kim et al. suggest two reasons why telicity is not a significant predictor in any of their models. First, telicity might be a token-level property that cannot be evaluated at the level of the verb type. In the literature on lexical aspect, some have argued that telicity is only ever calculated at the level of the event description (e.g., \citealt{Verkuyl72, Verkuyl93}), but others have maintained that verbs types in isolation do have an intrinsic telicity status, but that it can be modulated by the sentential context (e.g., \citealt{RothsteinEvents}). Second, telicity might simply be too complex a concept for subjects to evaluate.

To adjudicate between these two possibilities, we compared Kim et al.'s data to ratings from \citet{GanttDecomposing}, who asked subjects to annotate entire events as opposed to verbs in isolation for telicity. The questions used were similar (Kim et al.: \textit{Some verbs refer to an activity that could continue for an indefinite period of time, whereas other verbs refer to the completion of an event. To what extent does this verb refer to an event with a defined state of completion?}; Gantt et al.: \textit{Does the event have a natural endpoint?}). Annotator confidence ratings in Gantt et al.'s study are very high overall, but 39.3\% of lemmas receive different telicity ratings in different contexts.\footnote{We excluded from this count tokens with a telicity confidence rating of 0, 1, 2 on a scale from 0 to 4, keeping only highly confident labels.} This suggests that the primary source of noise in the telicity ratings is that subjects had difficulties rating verbs in isolation.

We suggest that our seed-based measure of telicity, while equally computed for verb types, circumvents this problem to some extent since the embeddings it operates on average over different usages of the verb. It is an open question why speakers do not seem to be able to intuit the average telicity status of a verb type in a similar fashion. In the conclusion of this paper, we discuss how the effect of telicity on syntax could be determined more conclusively using token embeddings.

\paragraph{Agentivity}

Table \ref{loo_ag} shows the results from the LOO comparison using only agentivity predictors. Compared to telicity, the difference between the ELPD improvements resulting from the seeds model and the ratings model, respectively, is less pronounced (466.5 vs. 147.4 for agentivity, 575.2 vs. 69.2 for telicity). To better understand where seeds and ratings diverge, we performed a pointwise comparison by subtracting the seed-based predictors from the ratings. Table \ref{highlow} shows the verbs for which the measures diverge most such that seed scores are high, but ratings low.

\begin{table}[h!]
    \centering
    \begin{tabular}{l c c}
       models  &  ELPD diff & SD \\
         \hline 
       \rule{0pt}{12pt}seeds vs. null model & 466.5 & 29.6\\
       ratings vs. null model & 147.4 & 17.5\\
       seeds+ratings vs. seeds & 22.1 & 7.0 \\
      seeds+ratings vs. ratings & 341.2 & 25.5\\
    \end{tabular}
    \caption{LOO analysis with only agentivity predictors.}
    \label{loo_ag}
\end{table}

\begin{table}[h!]
    \centering
    \begin{tabular}{l c c }
       verb  & seed score & rating \\
        \hline 
       \rule{0pt}{12pt}\textit{snore} & .645 & .42 \\
       \textit{stumble} & .371 &  .55 \\
       \textit{twinkle} & .369 & .69 \\
       \textit{tremble} & .321 & .67 \\
       \textit{stink} & .455 & .97 \\
       \textit{shiver} & .078 & .7 \\
      \textit{sneeze} & .237 & .89 \\
    \end{tabular}
    \caption{Verbs with high seed scores and low ratings.}
    \label{highlow}
\end{table}

Kim et al. operationalized agentivity by asking: \textit{To what extent does this verb describe something that is actively or intentionally done?} Accordingly, the verbs in Table \ref{highlow} denote activities that are not done intentionally or purposefully. The reason why the seed-based measure nonetheless rates them comparatively high for agentivity appears to be that most of them predominantly take human subjects, considering that the seed words contain several pronouns -- \textit{you}, \textit{he}, \textit{she} -- that can only refer to human participants. That is, the agentivity dimension appears to be sensitive to a property like animacy or sentience.

We tested this prediction using data from \citet{VanArsdall}, who collected ratings for 1,200 concrete nouns on 6 different animacy dimensions: general living/non-living scale, ability to think, ability to reproduce, similarity to a person, goal-directedness and movement likelihood. Using factor analysis, they further clustered these axes into two coarser dimensions, mental and physical animacy. Since all ratings were done on nouns, we fitted dimensions to these ratings using the new ranking loss function and then obtained values for Kim et al.'s intransitive verbs by projecting them onto these dimensions.

Table \ref{crossvalidation} summarizes the results of a 5-fold cross-validation, indicating a solid fit to the human ratings. In addition, since dimensions fitted to noun ratings might not work well for verbs, we computed the corelation between the 8 animacy axes and Kim et al.'s agentivity ratings, for which subjects were asked how intentionally an action was performed. As expected, we find a significant correlation for Thought (.191, \textit{p}-val. .001), Person (.191, \textit{p}-val. .001), Goals (.286, \textit{p}-val. .0) and coarse-grained Mental Animacy (.254, \textit{p}-val. .0), but not for the Physical Animacy features. This signals that while the animacy axes were fitted to noun ratings, they carry over to verbs.

\begin{table}[h!]
    \centering
        \begin{tabular}{l c c c}
        feature &  poc & xpoc & Pearson R \\
         \hline 
        \rule{0pt}{12pt}Living 	& .791 & 	.792  	& .788\\
        Thought &	.780 &	.785 	 	& .809\\
        Repr. 	& .789 & 	.791 	 	& .800\\
        Person & 	.779 	& .784 	& .801\\
        Goals 	& .796 & 	.800 		& .800\\
        Move 	& .764 & 	.767 	 &	.740\\
        Mental & 	.811 	& .813 	& .808\\
        Physical 	& .794 &	.795  	& .808\\
    \end{tabular}
    \caption{Crossvalidating the fitted dimensions on the \citet{VanArsdall} data.}
    \label{crossvalidation}
\end{table}

Next, we compute how much each of the 8 animacy axes improve the ELPD compared to a null model (Table \ref{loo_anim}). The strongest improvement is achieved by Move (444.6), which performs almost as well as the agentivity seeds (466.5). Goals has the weakest effect; Physical Animacy is stronger than Mental Animacy.

\begin{table}[h!]
    \centering
    \begin{tabular}{l c c }
        models &  ELPD diff & SD \\
         \hline 
        \rule{0pt}{12pt}Living vs. null model & 214.4 & 2.6\\
        Thought vs. null model & 195.5 & 19.2\\
        Reproduction vs. null model & 149.2 & 17.2\\
        Person vs. null model & 184.3 & 18.8\\
        Goals vs. null model & 74.3 & 12.1\\
        Move vs. null model & 444.6 & 28.5\\
        Mental vs. null model & 156.3 & 17.4\\
        Physical vs. null model & 23.5 & 21.2\\
    \end{tabular}
    \caption{LOO analysis with animacy axes in isolation.}
    \label{loo_anim}
\end{table}

We then compare how much each of the 8 axes improves on the agentivity seeds model and the agentivity ratings model, respectively (Table \ref{loo_anim2}). No telicity predictors were included. With the exception of Goals, all animacy axes improve the ratings model considerably more than the seeds model. The difference is more pronounced for physical than for mental animacy. 

\begin{table}[h!]
    \centering
    \begin{tabular}{l c c}
        models &  ELPD diff & SD \\
         \hline 
        \rule{0pt}{12pt}seeds+Living vs. seeds & 98.5 &14.4\\
        ratings+Living vs. ratings & 205.7 & 2.3\\
                 \hline 
        \rule{0pt}{12pt}seeds+Thought vs. seeds & 95.4 & 13.6\\
        ratings+Thought vs. ratings & 142.3 & 17.6\\
                 \hline 
        \rule{0pt}{12pt}seeds+Repr. vs. seeds & 78.6 & 12.7\\
        ratings+Repr. vs. ratings & 156.6 & 17.6\\
                 \hline 
        \rule{0pt}{12pt}seeds+Person vs. seeds & 64.1 & 11.3\\
        ratings+Person vs. ratings & 127.4 & 15.9\\
                 \hline 
        \rule{0pt}{12pt}seeds+Goals vs. seeds & 56.7 & 1.9\\
        ratings+Goals vs. ratings & 32.4 & 8.4\\
                 \hline 
        \rule{0pt}{12pt}seeds+Move vs. seeds & 192.6 & 19.4\\
        ratings+Move vs. ratings & 388.8 & 27.2\\
                 \hline 
        \rule{0pt}{12pt}seeds+Mental vs. seeds & 78.9 & 12.6\\
        ratings+Mental vs. ratings & 98.6 & 14.2\\
                 \hline 
        \rule{0pt}{12pt}seeds+Physical vs. seeds & 92.9 & 14.1\\
        ratings+Physical vs. ratings & 221.4 & 21.0\\
    \end{tabular}
    \caption{LOO analysis with animacy axes added to the seeds and ratings models.}
    \label{loo_anim2}
\end{table}

Therefore, we argue that the superior performance of the seeds compared to the ratings is due to the fact that they conceptualized agentivity differently. The ratings target specifically intentional, goal-oriented behavior, whereas the seeds target animacy construed more broadly. In particular, syntax appears to be more sensitive to features of purely physical animacy such as the ability to move than to volition and purpose. This also explains why potential seed words such as \textit{deliberate} and \textit{intentional} decreased the performance of the seeds model.

Finally, we note that both seeds and ratings fall short in that they cannot distinguish between true intransitives and their morphologically identical transitive variants. One verb for which even the seeds model makes poor predictions is \textit{break}, corresponding to the most extreme outlier in Figure 1 (observed mean rating: $\sim$ 4.5, predicted: $\sim$ 2.7). \textit{Break} scores relatively high on agentivity for either measure (rating: 3.24, seed score: .991) compared to, e.g., \textit{fall} (rating: .94, seed score: -.067). We attribute this to the fact that \textit{break} but not \textit{fall} has a transitive use, as in \textit{Ahmad broke the glass}. Recall that agentivity has been argued to matter for the syntax of intransitives in that in unergatives, the sole argument has the same syntactic status as the subject of transitives and equally receives an agent $\theta$-role. However, in \textit{Ahmad broke the glass}, it is the subject \textit{Ahmad} who receives an agent $\theta$-role, not the argument of the intransitive, \textit{the glass}. Hence, the fact that transitive \textit{break} is perceived as quite agentive does not signal an unergative syntax. As argued previously for telicity, evaluating agentivity at the level of the verb type comes with inherent limitations. 

\section{Conclusion}

In this paper, we have developed a novel approach to exploring the connection between unergativity/unaccusativity and agentivity/telicity. We set up interpretable dimensions using seed words associated with agentivity and telicity so as to best predict Kim et al.'s syntactic measure of unergativity/unaccusativity, the acceptability of a given verb in prenominal participle constructions. We then compared this model to human ratings for agentivity, telicity and animacy to understand why these dimensions predict the syntactic data so well. 

For agentivity, we have argued that the reason why the seeds outperform the ratings is that the latter construe agentivity too narrowly, in the sense of intentional, goal-directed behavior, while syntax appears to be sensitive to more low-level physical animacy features such as the ability to move. To reach this conclusion, we have also relied on a new method for fitting dimensions to human ratings that does not rely on seed words. For telicity, our conclusions have been more cautious, but we have found evidence that the poor predictive power of telicity ratings is due to the fact that speakers struggled to evaluate verb types for telicity, a problem that our seeds arguably circumvent to some extent. 

Nonetheless, both dimensions and ratings are limited in that they are computed for verb types. The natural solution would be to turn to token embeddings instead. If property axes for telicity and
agentivity could be established in token space, it
would be possible to determine directly if verb tokens occurring in unaccusative constructions like prenominal participles cluster around the telic and
non-agentive ends of these axes. However, as of yet,
we have no reliable method for computing interpretable dimensions for judging words in context. All previous semantic projection work that used token embeddings  targeted type-level judgments~\citep{lucy-etal-2022-discovering,ErkApidianikiAdjust, pmlr-v235-park24c,https://doi.org/10.1111/cogs.70072}, so there are no evaluations testing what techniques would work for token-level judgments. Moreover, of the papers that compared type-level and token-level input (again, for type-level judgments)~\citep{lucy-etal-2022-discovering,ErkApidianikiAdjust, https://doi.org/10.1111/cogs.70072}, only \citet{lucy-etal-2022-discovering} found improvements from token-level data. Methodological advances in this respect would be of value for many areas of research, including the syntax-semantics interface.

\bibliographystyle{acl_natbib}
\bibliography{references,additional}

\begin{thebibliography}{37}
\expandafter\ifx\csname natexlab\endcsname\relax\def\natexlab#1{#1}\fi

\bibitem[{Acart{\"u}rk and Zeyrek(2010)}]{AcarturkZeyrep}
Cengiz Acart{\"u}rk and Deniz Zeyrek. 2010.
\newblock Unaccusative/unergative distinction in {T}urkish: A connectionist
  approach.
\newblock In \emph{Proceedings of the 8\textsuperscript{th} Workshop on {A}sian
  Language Resources}, pages 111--119, Beijing, China.

\bibitem[{Allman(2017)}]{AllmannUnacc}
JungAe~Lee Allman. 2017.
\newblock \emph{Empirical examination of two diagnostics of {K}orean
  unaccusativity}.
\newblock Ph.D. thesis, The University of Texas at Arlington, Arlington, TX.

\bibitem[{Baker(2019)}]{BakerSplit}
James Baker. 2019.
\newblock \href {https://doi.org/10.1017/S1360674317000533} {Split
  intransitivity in {E}nglish}.
\newblock \emph{English Language and Linguistics}, 23:557–589.

\bibitem[{Binder et~al.(2016)Binder, Conant, Humphries, Fernandino, Simons,
  Aguilar, and Desai}]{BinderFeatures}
Jeffrey~R Binder, Lisa~L Conant, Colin~J Humphries, Leonardo Fernandino,
  Stephen~B Simons, Mario Aguilar, and Rutvik~H Desai. 2016.
\newblock \href {https://doi.org/10.1080/02643294.2016.1147426} {Toward a
  brain-based componential semantic representation}.
\newblock \emph{Cognitive Neuropsychology}, 33:130--174.

\bibitem[{Bolukbasi et~al.(2016)Bolukbasi, Chang, Zou, Saligrama, and
  Kalai}]{BolukbasiDim}
Tolga Bolukbasi, Kai-Wei Chang, James~Y Zou, Venkatesh Saligrama, and Adam~T
  Kalai. 2016.
\newblock Man is to computer programmer as woman is to homemaker? {D}ebiasing
  word embeddings.
\newblock \emph{Advances in neural information processing systems}, 29.

\bibitem[{Burzio(1981)}]{BurzioDiss}
Luigi Burzio. 1981.
\newblock \emph{Intranitive Verbs and {I}talian Auxiliaries}.
\newblock Ph.D. thesis, MIT, Cambridge, MA.

\bibitem[{Burzio(1986)}]{BurzioBook}
Luigi Burzio. 1986.
\newblock \href {https://doi.org/https://doi.org/10.1007/978-94-009-4522-7}
  {\emph{Italian Syntax: A {G}overnment and {B}inding Approach}}.
\newblock D. Reidel, Dordrecht.

\bibitem[{Carter et~al.(2025)Carter, Keller, and
  Hoffman}]{https://doi.org/10.1111/cogs.70072}
Georgia-Ann Carter, Frank Keller, and Paul Hoffman. 2025.
\newblock \href {https://doi.org/https://doi.org/10.1111/cogs.70072}
  {Leveraging context for perceptual prediction using word embeddings}.
\newblock \emph{Cognitive Science}, 49(6):e70072.

\bibitem[{Dowty(1991)}]{Dowty}
David Dowty. 1991.
\newblock \href {https://doi.org/10.2307/415037} {Thematic proto-roles and
  argument selection}.
\newblock \emph{Language}, 67:547--619.

\bibitem[{Engels et~al.(2025)Engels, Liao, Michaud, Gurnee, and
  Tegmark}]{Engels-nonlinear}
Joshua Engels, Isaac Liao, Eric~J Michaud, Wes Gurnee, and Max Tegmark. 2025.
\newblock Not all language model features are linear.
\newblock In \emph{Proceedings of ICLR}.

\bibitem[{Erk and Apidianaki(2024)}]{ErkApidianikiAdjust}
Katrin Erk and Marianna Apidianaki. 2024.
\newblock \href {https://doi.org/10.18653/v1/2024.naacl-long.146} {Adjusting
  interpretable dimensions in embedding space with human judgments}.
\newblock In \emph{Proceedings of the 2024 Conference of the North American
  Chapter of the Association for Computational Linguistics: Human Language
  Technologies (Volume 1: Long Papers)}, pages 2675--2686, Mexico City, Mexico.
  Association for Computational Linguistics.

\bibitem[{Gantt et~al.(2022)Gantt, Glass, and White}]{GanttDecomposing}
William Gantt, Lelia Glass, and Aaron~Steven White. 2022.
\newblock \href {https://doi.org/10.1162/tacl_a_00445} {Decomposing and
  recomposing event structure}.
\newblock \emph{Transactions of the Association for Computational Linguistics},
  10:17--34.

\bibitem[{Gar{\'i}~Soler and
  Apidianaki(2020)}]{gari-soler-apidianaki-2020-bert}
Aina Gar{\'i}~Soler and Marianna Apidianaki. 2020.
\newblock {BERT} knows {P}unta {C}ana is not just beautiful, it{'}s gorgeous:
  Ranking scalar adjectives with contextualised representations.
\newblock In \emph{Proceedings of EMNLP}, pages 7371--7385.

\bibitem[{Graf et~al.(2017)Graf, Philipp, Xu, Kretzschmar, and Primus}]{Graf}
Tim Graf, Markus Philipp, Xiaonan Xu, Franziska Kretzschmar, and Beatrice
  Primus. 2017.
\newblock \href {https://doi.org/10.1016/j.lingua.2017.08.006} {The interaction
  between telicity and agentivity: Experimental evidence from intransitive
  verbs in {G}erman and {C}hinese}.
\newblock \emph{Lingua}, 200:84--106.

\bibitem[{Grand et~al.(2022)Grand, Blank, Pereira, , and Fedorenko}]{GrandDim}
Gabriel Grand, Idan~Asher Blank, Francisco Pereira, , and Evelina Fedorenko.
  2022.
\newblock \href {https://doi.org/10.1038/s41562-022-01316-8} {Semantic
  projection recovers rich human knowledge of multiple object features from
  word embeddings}.
\newblock \emph{Nature Human Behaviour}, 6:975--987.

\bibitem[{Huang(2018)}]{HuangUnaccusativity}
Yujing Huang. 2018.
\newblock \emph{Linking form to meaning: {R}eevaluating the evidence for the
  unaccusative hypothesis}.
\newblock Ph.D. thesis, Harvard University, Cambridge, MA.

\bibitem[{Jackendoff(1983)}]{Jackendoff83}
Ray Jackendoff. 1983.
\newblock \emph{Semantics and Cognition}.
\newblock MIT Press, Cambridge, MA.

\bibitem[{Kim et~al.(2024)Kim, Binder, Humphries, and Conant}]{KimDecomposing}
Songhee Kim, Jeffrey~R Binder, Colin Humphries, and Lisa~L Conant. 2024.
\newblock \href {https://doi.org/10.1080/23273798.2024.2368119} {Decomposing
  unaccusativity: a statistical modelling approach}.
\newblock \emph{Language, Cognition and Neuroscience}, 39:1189--1211.

\bibitem[{Kozlowski et~al.(2019)Kozlowski, Taddy, and Evans}]{KozlowskiDim}
Austin~C. Kozlowski, Matt Taddy, and James~A. Evans. 2019.
\newblock \href {https://doi.org/10.1177/0003122419877135} {The geometry of
  culture: {A}nalyzing the meanings of class through word embeddings}.
\newblock \emph{American Sociological Review}, 84:905--949.

\bibitem[{Levin(1993)}]{LevinEnglish}
Beth Levin. 1993.
\newblock \emph{English verb classes and alternations: {A} preliminary
  investigation}.
\newblock University of Chicago Press.

\bibitem[{Levin and {Rappaport Hovav}(1995)}]{LRH}
Beth Levin and Malka {Rappaport Hovav}. 1995.
\newblock \emph{Unaccusativity. At the Syntax-Lexical Semantics Interface}.
\newblock MIT Press, Cambridge, MA.

\bibitem[{Lucy et~al.(2022)Lucy, Tadimeti, and
  Bamman}]{lucy-etal-2022-discovering}
Li~Lucy, Divya Tadimeti, and David Bamman. 2022.
\newblock Discovering differences in the representation of people using
  contextualized semantic axes.
\newblock In \emph{Proceedings of EMNLP}, pages 3477--3494.

\bibitem[{Nayyeri et~al.(2019)Nayyeri, Zhou, Vahdati, Yazdi, and
  Lehmann}]{nayyeri2019adaptivemarginrankingloss}
Mojtaba Nayyeri, Xiaotian Zhou, Sahar Vahdati, Hamed~Shariat Yazdi, and Jens
  Lehmann. 2019.
\newblock \href {http://arxiv.org/abs/1907.05336} {Adaptive margin ranking loss
  for knowledge graph embeddings via a correntropy objective function}.

\bibitem[{Park et~al.(2024)Park, Choe, and Veitch}]{pmlr-v235-park24c}
Kiho Park, Yo~Joong Choe, and Victor Veitch. 2024.
\newblock The linear representation hypothesis and the geometry of large
  language models.
\newblock In \emph{Proceedings of ICML}, pages 39643--39666.

\bibitem[{Perlmutter(1978)}]{Perlmutter}
David Perlmutter. 1978.
\newblock \href {https://doi.org/10.3765/bls.v4i0.2198} {Impersonal passives
  and the {U}naccusative {H}ypothesis}.
\newblock \emph{Papers from the Annual Meeting of the Berkeley Linguistic
  Society}, 4:157--189.

\bibitem[{Rothstein(2008)}]{RothsteinEvents}
Susan Rothstein. 2008.
\newblock \href {https://doi.org/10.1002/9780470759127} {\emph{Structuring
  events: {A} study in the semantics of lexical aspect}}.
\newblock Blackwell, Oxford.

\bibitem[{Sorace(2000)}]{SoraceLang}
Antonella Sorace. 2000.
\newblock \href {https://doi.org/10.2307/417202} {Gradients in auxiliary
  selection with intransitive verbs}.
\newblock \emph{Language}, 76(4):859--890.

\bibitem[{Sorace(2004)}]{SoracePuzzle}
Antonella Sorace. 2004.
\newblock \href {https://doi.org/10.1093/acprof:oso/9780199257652.003.0010}
  {Gradience at the lexicon-syntax interface: {E}vidence from auxiliary
  selection and implications for unaccusativity}.
\newblock In Artemis Alexiadou, Elena Anagnostopoulou, and Martin Everaert,
  editors, \emph{The Unaccusativity Puzzle}, pages 243--268. Oxford UP, Oxford.

\bibitem[{Sorace(2011)}]{SoraceArchivio}
Antonella Sorace. 2011.
\newblock Gradience in split intransitivity: the end of the {U}naccusative
  {H}ypothesis?
\newblock \emph{Archivio Glottologico Italiano}, XCVI(1):67--86.

\bibitem[{Staub(2021)}]{StaubTestRetest}
Adrian Staub. 2021.
\newblock \href {https://doi.org/10.1016/j.jml.2020.104190} {How reliable are
  individual differences in eye movements in reading?}
\newblock \emph{Journal of Memory and Language}, 116:104190.

\bibitem[{Tenny(1987)}]{TennyDiss}
Carol Tenny. 1987.
\newblock \emph{Grammaticalizing Aspect and Affectedness}.
\newblock Ph.D. thesis, Cambridge, MA.

\bibitem[{Tenny(1994)}]{TennyBook}
Carol Tenny. 1994.
\newblock \href {https://doi.org/10.1007/978-94-011-1150-8} {\emph{Aspectual
  Roles and the Syntax-Semantics Interface}}.
\newblock Kluwer, Dordrecht.

\bibitem[{Van Valin~Jr(1990)}]{VanValin}
Robert~D Van Valin~Jr. 1990.
\newblock \href {https://doi.org/10.2307/414886} {Semantic parameters of split
  intransitivity}.
\newblock \emph{Language}, 66:221--260.

\bibitem[{VanArsdall and Blunt(2022)}]{VanArsdall}
Joshua~E VanArsdall and Janell~R Blunt. 2022.
\newblock \href {https://doi.org/10.3758/s13421-021-01266-y} {Analyzing the
  structure of animacy: {E}xploring relationships among six new animacy and 15
  existing normative dimensions for 1,200 concrete nouns}.
\newblock \emph{Memory \& Cognition}, 50:997--1012.

\bibitem[{Vendler(1957)}]{Vendler}
Zeno Vendler. 1957.
\newblock \href {https://doi.org/10.2307/2182371} {Verbs and times}.
\newblock \emph{The Philosophical Review}, 66(2):143--160.

\bibitem[{Verkuyl(1972)}]{Verkuyl72}
Henk~Jacob Verkuyl. 1972.
\newblock \emph{On the Compositional Nature of the Aspects}.
\newblock Kluwer, Dordrecht.

\bibitem[{Verkuyl(1993)}]{Verkuyl93}
Henk~Jacob Verkuyl. 1993.
\newblock \emph{A Theory of Aspectuality}.
\newblock Cambridge UP, Cambridge.

\end{thebibliography}


\end{document}